# Using Artificial Bee Colony Algorithm for MLP Training on Earthquake Time Series Data Prediction

Habib Shah, Rozaida Ghazali, and Nazri Mohd Nawi

**Abstract**—Nowadays, computer scientists have shown the interest in the study of social insect's behaviour in neural networks area for solving different combinatorial and statistical problems. Chief among these is the Artificial Bee Colony (ABC) algorithm. This paper investigates the use of ABC algorithm that simulates the intelligent foraging behaviour of a honey bee swarm. Multilayer Perceptron (MLP) trained with the standard back propagation algorithm normally utilises computationally intensive training algorithms. One of the crucial problems with the backpropagation (BP) algorithm is that it can sometimes yield the networks with suboptimal weights because of the presence of many local optima in the solution space. To overcome ABC algorithm used in this work to train MLP learning the complex behaviour of earthquake time series data trained by BP, the performance of MLP-ABC is benchmarked against MLP training with the standard BP. The experimental result shows that MLP-ABC performance is better than MLP-BP for time series data.

——————————— ◆ ———————————

## 1 INTRODUCTION

ARTIFICIAL Neural Networks (ANNs) are the most novel and powerful artificial tool suitable for solving combinatorial problems such as prediction and classification [1], [2], [3], [4]. ANNs are being used extensively for solving universal problems intelligently like continuous, discrete, and clustering [5], [6]. ANNs are being applied for different optimisation and mathematical problems such as classification, object and image recognition, signal processing, seismic events prediction, temperature and weather forecasting, bankruptcy, tsunami intensity, earthquake, and sea level [5], [7], [8], [9], [10]. Different techniques are used for optimal network performance for training ANNs such as evolutionary algorithms (EA), genetic algorithms (GA), partial swarm optimisation (PSO), differential evolution (DE), ant colony, and backpropagation algorithm [11], [12], [13], [14], [15], [16], [17]. These techniques are used for initialisation of optimal weights, parameters,

———————————————

- H. Shah is with the Universiti Tun Hussein Onn Malaysia, 86400 Parit Raja, Batu Pahat. E-mail: habibshah.uthm@gmail.com.
- R. Ghazali with the Universiti Tun Hussein Onn Malaysia, 86400 Parit Raja, Batu Pahat. E-mail: rozaida@uthm.edu.my.
- N Mohd Nawi with the Universiti Tun Hussein Onn Malaysia, 86400 Parit Raja, Batu Pahat. E-mail: nazri@uthm.edu.my.

activation function, and selection of proper network structure [18].

Backpropagation (BP) algorithm is accepted algorithm used for MLP training [19]. The main task of BP algorithm is to update the network weights for minimising output error using backpropagation processing because the accuracy of any approximation depends on the selection of proper weights for the neural networks (NNs). It has high success rate in solving many complex problems but it still has some drawbacks especially when setting parameter values like initial values of connection weights, value for learning rate, and momentum. If the network topology is not carefully selected, the NNs algorithm can get trapped in local minima or it might lead to slow convergence or even network failure. In order to overcome the disadvantages of standard BP, many global optimisation population-based techniques have been proposed for MLP training such as EA, GA, improved GA, DE, BP-ant colony, and PSO [11], [12], [13], [16], [17].

Artificial Bee Colony (ABC) algorithm is a population-based algorithm that can provide the best possible solutions for different mathematical problems by using inspiration techniques from nature [20]. A common feature of population-based algorithms is that the population consisting of feasible solutions to the difficulty is customised by applying some agents on the solutions depending on the information of their robustness. Therefore, the population is encouraged towards improved solution areas of the solution space. Population-based optimisation algorithms are categorised into two sections namely evolutionary algorithm (EA) and SI-based algorithm [21], [22]. In EA, the major plan underlying this combination is to take the weight matrices of the ANNs as individuals, to change the weights by means of some operations such as crossover and mutation, and to use the error produced by the ANNs as the fitness measure that guides selection. In SI-based algorithm, ABC has the advantage of global optimisation and easy recognition. It has been successfully used in solving combinatorial optimisation problems such as clustering and MLP training for XOR problem [23], [24]. This is one of the self-organising and highly approachable solutions for different computational and mathematical problems. ABC algorithm is an easily understandable technique for training MLP on classification problems [25]. This algorithm uses randomly selected natural techniques with colony to train NNs by optimal weights [26]. In this study, ABC algorithm is used successfully to train MLP on earthquake time series data for prediction task. The performance of the algorithm is compared with standard BP algorithm.

This paper is organised as follows: A brief review on ANN and ABC and BP algorithms is given in Section 2 and Section 3, respectively. The proposed ABC algorithm and the training and testing of the network using ABC algorithm are detailed in Section 4. Section 5 contains the prediction of earthquake event. Results and discussion are discussed in Section 6. Finally, the paper is concluded in Section 7.

## 2 ARTIFICIAL NEURAL NETWORKS

### 2.1 TRAINING OF MLP NEURAL NETWORKS

MLP was introduced in 1957 to solve different combinatorial problems [27]. MLP, which is also known as feed forward neural networks was first introduced for the non-linear XOR, and was then successfully applied to different combinatorial problems. MLP is mostly used for information processing and pattern recognition in prediction of seismic activities. In this section, MLP's characteristics and interaction with the seismic signals are explained. MLP works as a universal approximation in which inputs signal propagates in forward direction. It is highly used and tested with different problems such as in time series prediction and function approximation [1], [3], [4], [8]. Figure 1 shows the architecture of MLP with two hidden layers, one output layer, and one input layer.

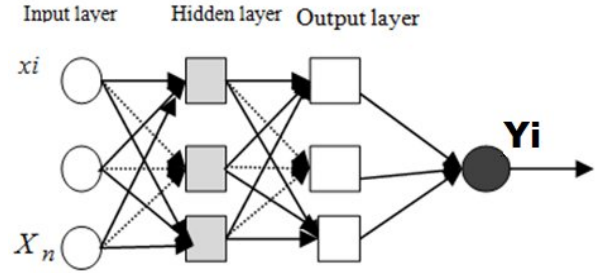

Fig 1: Multi Layer Perceptron Neural Network

$$Y_i = f_i \left( \sum_{j=1}^{n} w_{ij} x_j + b_i \right) \quad (1)$$

where $y_i$ is the output of the node, $x_i$ is the jth input to the node, $w_{ij}$ is the connection weight between the input node and output node, $\theta_i$ is the threshold (or bias) of the node, and $f_i$ is the node transfer function. Usually, the node transfer function is a non-linear function such as a sigmoid function, a Gaussian function, and etc. The network error function E will be minimised as

$$E(w(t)) = \frac{1}{n} \sum_{j=1}^{n} \sum_{k=1}^{k} (d_k - O_t) \quad (2)$$

where E (w (t)) is the error at the tth iteration; w(t) is the weights in the connections at the tth iteration; dk is the desired output node; ok is the actual value of the kth output node; K is the number of output nodes; and n is the number of patterns. T is the optimisation target to minimise the objective function by optimising the network weights w(t).

## 3 ARTIFICIAL BEE COLONY

### 3.1 Swarm Intelligence

Since the last two decades, swarm intelligence (SI) has been the focus of many researches because of its unique behaviour inherent from the social insects [13], [14], [22], [25]. Bonabeau has defined the SI as "any attempt to design algorithm or distributed problem-solving devices inspired by the collective behaviour of social insect colonies and other animal societies" [28]. He mainly focused on the behaviour of social insects alone such as termites, bees, wasps, and different ant species. However, swarm can be considered as any collection of interacting agents or individuals. Ants are individual agents of ACO [29]. An immune system can be considered as a group of cells and molecules as well as a crowd is a swarm of people [31]. PSO and ABC are popular population-based stochastic optimisation algorithms adapted for the optimisation of non-linear functions in multidimensional space [32].

## 3.2 Artificial Bee Colony algorithm

Artificial Bee Colony algorithm (ABC) was proposed for optimisation, classification, and NNs problem solution based on the intelligent foraging behaviour of honey bee swarm [20], [21], [22], [26]. Therefore, ABC is more successful and most robust on multimodal functions included in the set with respect to DE, PSO, and GA [16], [21], [23]. ABC algorithm provides solution in organised form by dividing the bee objects into different tasks such as employed bees, onlooker bees, and scout bees. These three bees/tasks determine the objects of problems by sharing information to others bees. The common duties of these artificial bees are as follows:

Employed bees: Employed bees use multidirectional search space for food source with initialisation of the area. They get information and all possibilities to find food source and solution space. Sharing of information with onlooker bees is performed by employee bees. An employed bee produces a modification on the source position in her memory and discovers a new food source position. Provided that the nectar amount of the new source is higher than that of the previous source, the employed bee memorizes the new source position and forgets the old one.

Onlooker bees: Onlooker bees evaluate the nectar amount obtained by employed bees and choose a food source depending on the probability values calculated using the fitness values. For this purpose, a fitness-based selection technique can be used. Onlooker bees watch the dance of hive bees and select the best food source according to the probability proportional to the quality of that food source.

Scout bees: Scout bees select the food source randomly without experience. If the nectar amount of a food source is higher than that of the previous source in their memory, they memorise the new position and forget the previous position. Whenever employed bees get a food source and use the food source very well again, they become scout bees to find new food source by memorising the best path. The detailed pseudocode of ABC algorithm is shown as follows:

1: Initialise the population of solutions $X_i$ where i=1…..SN
2: Evaluate the population
3: Cycle=1
4: Repeat from step 2 to step 13
5: Produce new solutions (food source positions) $V_{i,j}$ in the neighbourhood of $x_{i,j}$ for the employed bees using the formula

$$V_{i,j} = x_{i,j} + \Phi_{ij}(x_{i,j} - x_{k,j}) \quad (3)$$

where k is a solution in the neighbourhood of i, $\Phi$ is a random number in the range [-1, 1] and evaluate them.
6: Apply the Greedy Selection process between process
7: Calculate the probability values $p_i$ for the solutions $x_i$ by means of their fitness values by using formula

$$p_i = \frac{fit_i}{\sum_{k=1}^{S.N} fit_n} \quad (4)$$

The calculation of fitness values of solutions is defined as

$$fit_i = \begin{cases} \frac{1}{1+f_i} & f_i >= 0 \\ 1+abs(f_i) & f_i < 0 \end{cases} \quad (5)$$

Normalise $p_i$ values into [0, 1]
8: Produce the new solutions (new positions) $v_i$ for the onlookers from the solutions $x_i$, selected depending on $P_i$, and evaluate them
9: Apply the Greedy Selection process for the onlookers between $x_i$ and $v_i$
10: Determine the abandoned solution (source), if exists, replace it with a new randomly produced solution $x_i$ for the scout using the following equation

$$x_i^j = x_{min}^j + rand(0,1)(x_{max}^j - x_{min}^j) \quad (6)$$

11: Memorise the best food source position (solution) achieved so far
12: cycle=cycle+1
13: until cycle= Maximum Cycle Number (MCN)

## 4 THE PROPOSED FRAMEWORK FOR MLP-ABC

The proposed flowchart of the ABC algorithm for earthquake time series data prediction is given in Figure 2. In the figure, each cycle of the search consists of three steps after initialisation of the colony, foods, and three control parameters in the number of food sources, which are equal to the number of employed bees or onlooker bees (SN), the value of limit, the maximum cycle number (MCN) for MLP-ABC algorithm. The initialisation of weights was compared with output and the best weight cycle was selected by scout bees' phase. The bees (employed bees, onlooker bees) would continue searching until the last cycle to find the best weights for networks. The food source of which the nectar was neglected by the bees was replaced with a new food source by the scout bees. Every bee (employed bees, onlooker bees) would produce new solution area for the network and the Greedy Selection would decide the best food source position. Suppose that the neglected source is $x_i$ and $j \in \{1, 2... D\}$, then the scout bees determined a new food source to be replaced with $x_i$.

The foods area was limited in range [10,-10]. It was applied randomly and was initialised for evaluation. This operation can be defined by using equation 6. Every bee (employed bees, onlooker bees) would produce new evaluated solution area for the network and the Greedy Selection was decided for the best food source position. If the new food source has equal or better nectar than the old food source, it was replaced with the new food source in the memory. Otherwise, the old food source was retained in the memory. The basic idea of ABC scheme is to use agents of bees to search the best combination of weights for network. All steps in finding optimal weights for network are shown in proposed ABC algorithm framework in Figure 2. The figure shows how to find and select the best weights and how to replace with the previous one. The Greedy Selection was applied between two sets of values $x_i$ and $V_i$ while the best scout bees were randomly selected.

The proposed frameworks can easily train earthquake time series data for prediction task by finding optimal network weights for MLP.

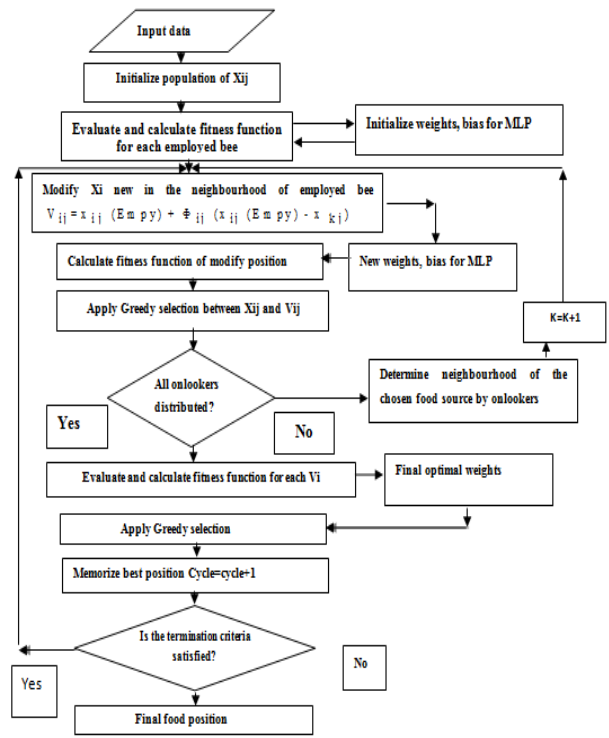

Fig 2: Proposed Flowchart for ABC algorithm to train MLP by ABC

## 5 PREDICTION OF EARTHQUAKE EVENT

In this research, real-time series data of seismic event earthquake was selected for training and testing. The data from the Southern California Earthquake Data Center (SCEDC) holdings for 2011 were selected [30]. The data included local, regional, and quarry-blast events with epicentres between latitudes 32.0S and 37.0N and longitudes between -122.0W and -114.0E. There were four main earthquake parameters, namely depth of earthquake, time of occurrence, geographical area, and magnitude of earthquakes. The significant parameter earthquake magnitude by Richter scale was used for simulation of earthquake magnitude prediction. Data obtained from the SCEC website were used to define input classes and test the MLP-ABC model proposed in this research. The earthquake record of Southern California between 1st January 2010 and 30th May 2011 was divided into fifty data sets per day. The networks were tested for the prediction of earthquake magnitude from horizon one to horizon five by MLP-ABC and MLP-BP.

Neural networks have been used successfully to solve complicated pattern recognition and classification problems in different domains such as satellite data, GPS data, and financial forecasting [3], [34]. Recently, NNs are also applied for earthquake prediction by using different models such as Backpropagation Neural Networks (BPNNs), Radial-Basis Function (RBF) NNs, Recurrent NNs, and probabilistic NNs [1], [35], [36], [37], [38], [39], [40]. These models mostly use the seismicity indicators as the parameters. These models are limited to predict the earthquake magnitude of more than 7.5. Therefore, MLP-ABC can predict the magnitude of more than 7.5.

## 6 SIMULATION RESULTS

In order to evaluate the performance of the proposed ABC to train MLP for benchmark earthquake time series data scheme for prediction techniques, simulation experiments were performed on a 1.66 GHz Core 2 Duo Intel Workstation with 1GB RAM using Matlab 2010a software. The comparison of standard BP training and ABC algorithms is discussed based on the simulation results implemented in Matlab 2010a. The California earthquake data for the year 2011 were used taken from http://www.data.scec.org/. The earthquake parameter magnitude was used to train the MLP using ABC algorithm. The data were divided into two datasets: 70% for training and 30% for testing. The learning rate (LR) and momentum (MOM) were set to 0.6 and 0.5, respectively. It should to be noted that the range of weights are different for both experimentations where [1,-1] is for MLP-BP and [10,-10] is for MLP-ABC. The weight values of MLP-ABC were initialised, evaluated, and fitted using ABC algorithm, while the weight values of MLP-BP were adjusted from the range [1,-1] randomly. All the simulation parameters were taken as given in Table 1. Besides that, the minimum value of mean square errors (MSE) was selected for testing. The stopping criteria of minimum error were set to 0.0001 for MLP-BP while MLP-ABC was stopped on MCN. The MLP was trained with inputs, hidden, and an output node varying from 2 to 4, respectively.

During the experiment, 5 trials were performed for training MLP-ABC. Each case and run was started with different number of parameters and with random population of foods. The sigmoid function was used as activation function for network output. The value of "limit" is equal to FoodNumber × D where D is the dimension of the problem and FoodNumber is half of the colony size, which is 50.

When the number of input, hidden, and output nodes of the neural network and running time varied, the performance of ABC was stable, which is important for the designation of neural networks in the current state where there are no specific rules for the decision of the number of hidden nodes.

Finally, mean square errors (MSE) and normalised mean square error (NMSE) were calculated for MLP-BP and MLP-ABC algorithms. The simulation results showed the effectiveness and efficiency of ABC algorithm. The comparison simulation of different network structures is presented in Table 1. The network parameter, MCN, objective function evaluation (OFE), runtimes, network shape, and epochs are presented in Table 1.

TABLE 1
NETWORK PARAMETER FOR MLP-BP AND MLP-ABC

| Parameters | MLP-BP | MLP-ABC |
|---|---|---|
| LR | 0.6 | ———— |
| Momentum | 0.5 | ———— |
| Dimension | | From 6 to 28 |
| MCN | ———— | From 100 to 1000 |
| Epochs/OFE | 1000 to 3000 | 1000-5000 |
| No. of hidden nodes | From 2 to 4 nodes | |
| No. of inputs nodes | From 2 to 4 nodes | |
| No. of output nodes | From 1 to 4 nodes | |
| Weights range | [1,-1] | [10,-10] |
| Runtime | ———— | From 2 to 10 |

TABLE 2
AVERAGE RESULTS OF MLP-BP AND MLP-ABC FOR PREDICTION.

| Network Structure/MSE | MLP-ABC | MLP-BP |
|---|---|---|
| 2-2-1 | 0.00161368 | 0.0195048 |
| 2-3-1 | 0.00170239 | 0.0184944 |
| 3-3-1 | 0.00161061 | 0.0174701 |
| 4-2-4 | 0.00163705 | 0.0193873 |
| 4-4-2 | 0.00187162 | 0.0220181 |

where OFE=MCN × FoodSource.

The dimension (D) can change the structure of MLP-ABC network model selected where 6, 9, 13, 16, 22, and 28 showed 2-2-1, 2-3-1, 3-3-1, 4-2-4 and 4-4-4, respectively.

From Table 1, we can see that the maximum cycle numbers are less than maximum epochs for MLP-ABC training while the OFE increases. The network structure employed in this experimentation started from two inputs, two hidden, and one output layer up to four inputs, four hidden, and two outputs nodes that contained the MSE and NMSE for training MLP-ABC and MLP-BP algorithm on earthquake data.

The NMSE can be found by the following formulae:

$$NMSE = \frac{1}{\sigma_n^2} \sum_{i=1}^{n} (Y_i - \hat{Y}_i)^2 \qquad (7)$$

$$\sigma^2 = \frac{1}{n-1} \sum_{i=1}^{n} (Y_i - \overline{Y})^2 \qquad (8)$$

$$\overline{Y} = \frac{1}{n} \sum_{i=1}^{n} Y_i \qquad (9)$$

where n is the total number of given data, Y is the actual value of earthquake magnitude, and $\hat{Y}$ represents values of predicted magnitude of earthquake.

From Table 2, we can see the best result by 3-3-1 network structure. It can also be easily seen that the proposed ABC algorithm outperforms MLP-BP algorithm for training earthquake time series data with an NMSE of 0.01364948 whereas MLP-BP falls behind with an NMSE of .00241384 for 3-3-1. The process of convergence to global minima can be seen in Figure 3 and Figure 4 for MLP-ABC and MLP-BP, respectively. The best results of 3-3-1 network structure for training and testing of MLP-BP and MLP-ABC are given in Figures 5, 6, 7, and 8, respectively.

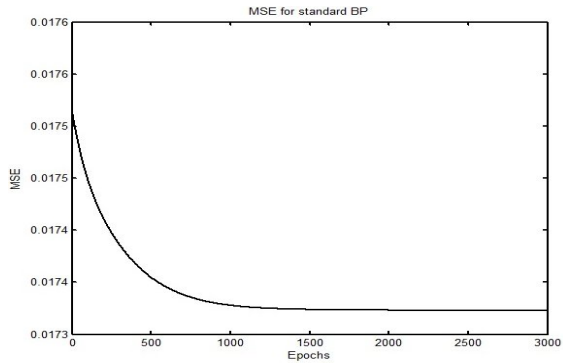

Fig 3: MSE of MLP-BP for Earthquake Data

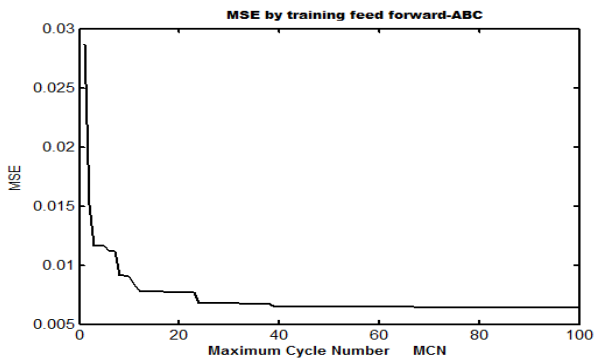

Fig 4: MSE by MLP-ABC for Earthquake Data

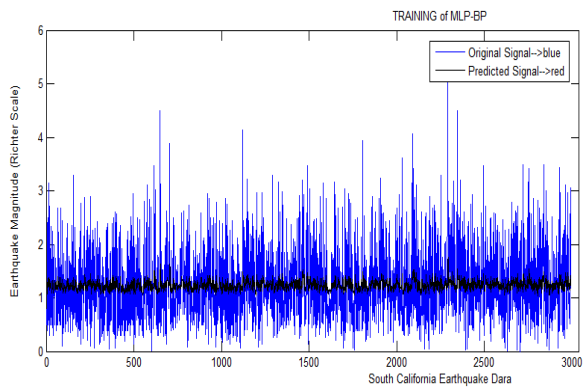

Fig 5: Training of Earthquake Data by MLP-BP

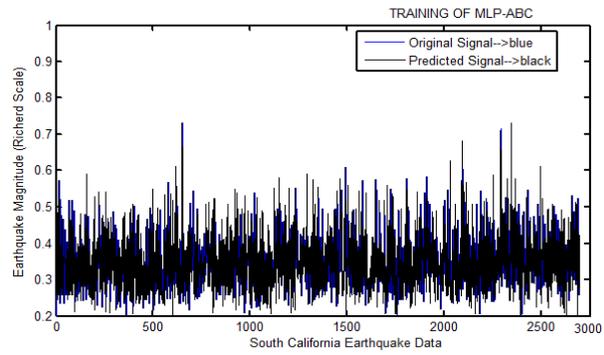

Fig 6: Training of Earthquake Data by MLP-ABC

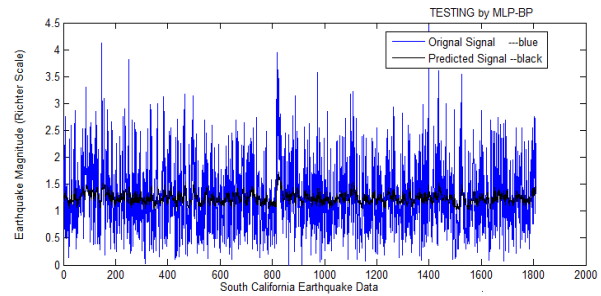

Fig 7: Testing of Earthquake Data by MLP-BP

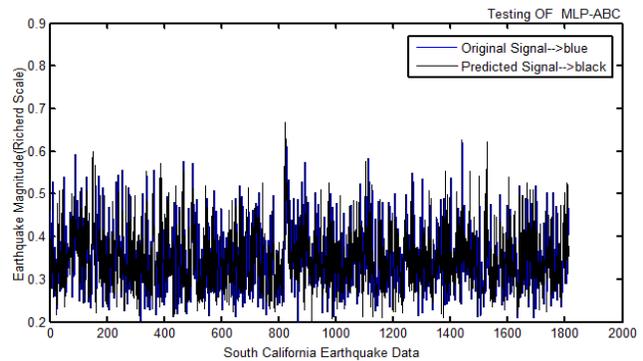

Fig 8: Testing of Earthquake Data by MLP-ABC

It can be seen in Figures 5, 6, 7, and 8 that ABC algorithm prediction follows the actual trend during the training and testing phases. Overall, ABC algorithm has shown 99.89 percent accuracy on time series earth quake data, which is significantly higher than MLP-BP algorithm. Meanwhile, the MLP-BP shows 89% accuracy, which is significantly lower than MLP-ABC.

## 7 CONCLUSION

The ABC algorithm combines the exploration and exploitation processes successfully, which proves the high performance of training MLP for earthquake time series data prediction. It has the powerful ability of searching global optimal solution. So, the proper weights of the algorithms may speed up the initialisation and improve the prediction accuracy of the trained NNs. The simulation results show that the proposed ABC algorithm can successfully train real-time data for prediction purpose, which further extends the quality of the given approach. The performance of ABC is compared with the traditional BP algorithm. ABC shows significantly higher results than backpropagation during experiment. ABC also shows higher accuracy in prediction. The proposed frameworks have successfully predicted the magnitude of earthquake.

## ACKNOWLEDGEMENT

The authors would like to thank University Tun Hussein Onn Malaysia (UTHM) for supporting this research under the Postgraduate Incentive Research Grant Vote No .0739.

**Habib Shah** is a Ph.D student at Universiti Tun Hussein Onn Malaysia (UTHM) since 2010. His current research focuses on the optimization of Artificial Neural Networks using Swarm Intelligence Algorithms. He got his Masters in Computer Science from Federal Urdu University of Arts, Science and Technology, Karachi, Pakistan in 2007. And his Bachelors in Computer Science from University of Malakand, Pakistan in 2005.

**Rozaida Ghazali** is currently a Deputy Dean (Research and Development) at the Faculty of Information Technology and Multimedia, Universiti Tun Hussein Onn Malaysia (UTHM). She graduated with Ph.D. degree from the School of Computing and Mathematical Sciences at Liverpool John Moores University, United Kingdom in 2007, on the topic of Higher Order Neural Networks for Financial Time series Prediction. Earlier, in 2003 she completed her M.Sc. degree in Computer Science from Universiti Teknologi Malaysia (UTM). She received her B.Sc. (Hons) degree in Computer Science (Information System) from the Universiti Sains Malaysia (USM) in 1997. In 2001, Rozaida joined the academic staff in UTHM. Her research area includes neural networks, data mining, financial time series prediction, data analysis, physical time series forecasting, and fuzzy logic.

**Nazri Mohd Nawi** received his B.S.degree in Computer Science from University of Science Malaysia (USM), Penang, Malaysia. His M.Sc. degree in computer science was received from University of Technology Malaysia (UTM), Skudai,Johor, Malaysia. He received his Ph.D. degree in Mechanical Engineering department, Swansea University, Wales Swansea. He is currently a Associate professor in Software Engineering Department at Universiti Tun Hussein Onn Malaysia (UTHM). His research interests are in optimization, data-mining techniques and neural networks.